\documentclass{article}

\usepackage{microtype}
\usepackage{graphicx}
\usepackage{subcaption}
\usepackage{booktabs}
\usepackage{tikz}
\usetikzlibrary{positioning,arrows.meta}
\usepackage{pgfplots}
\pgfplotsset{compat=1.18}
\usepackage{xurl}

\usepackage{hyperref}

\usepackage[accepted]{icml2026}

\usepackage{amsmath}
\usepackage{amssymb}
\usepackage{mathtools}
\usepackage{amsthm}

\usepackage[capitalize,noabbrev]{cleveref}

\theoremstyle{plain}

\theoremstyle{definition}

\theoremstyle{remark}

\begin{document}

\twocolumn[
  \icmltitle{Speedrunning Tabular Foundation Model Pretraining}

  \icmlsetsymbol{equal}{*}

  \begin{icmlauthorlist}
    \icmlauthor{Salih Bora Öztürk}{ufr}
    \icmlauthor{Alexander Pfefferle}{ellis,ufr}
    \icmlauthor{Frank Hutter}{prior,ellis,ufr}
  \end{icmlauthorlist}

  \icmlaffiliation{ufr}{University of Freiburg, Freiburg im Breisgau, Germany}
  \icmlaffiliation{ellis}{ELLIS Institute Tübingen, Tübingen, Germany}
  \icmlaffiliation{prior}{Prior Labs, Freiburg im Breisgau, Germany}

  \icmlcorrespondingauthor{Salih Bora Öztürk}{oeztuers@cs.uni-freiburg.de}

  \icmlkeywords{Machine Learning, Tabular Data, TabPFN, Foundation Models for Structured Data Workshop, ICML}

  \vskip 0.3in
]

\printAffiliationsAndNotice{}

\begin{abstract}
  Pretraining cost is a major bottleneck for research on tabular foundation models, slowing the iteration cycle for new architectures, priors, and optimization ideas.
  Yet the community lacks a simple way to compare and accumulate pretraining speedups.
  We introduce a community speedrun for nanoTabPFN: contributors modify a single-file training script and compete to reach a fixed downstream ROC AUC target on subsampled TabArena using one NVIDIA L40S GPU.
  The current best record reaches the target in $0.92$ minutes, an $81\times$ speedup over the $74.32$-minute baseline while using $22\times$ fewer synthetic datasets.
  The speedrun format provides a simple protocol for the community to add, verify, and stack pretraining improvements, with the leaderboard open to contributions.
  Code and records are available at \url{https://github.com/borawhocodess/modded-nanotabpfn}.
\end{abstract}

\section{Introduction}
Structured data in tables is the bedrock of decision making~\citep{borisov2024dnntabular,vanbreugel2024position}.
Tabular foundation models (TFMs) are fundamentally changing the field, much as LLMs did for text~\citep{hollmann2023tabpfn,hollmann2025tabpfnv2,grinsztajn2026tabpfn25advancingstateart,qu2025tabicl,qu2026tabiclv2,ma2025tabdpt,zhang2025limix}.
However, pretraining foundation models is computationally expensive, taking hours to days even at small scale and slowing research iteration.
Considerable work has tackled the problem of speeding up LLM pretraining~\citep{moddednanogpt,geiping2023cramming,deepseekv3,parametergolf2026}, but to our knowledge no equivalent effort exists for TFMs.
We close this gap with a public speedrun built around nanoTabPFN~\citep{pfefferle2025nanotabpfn}, empirically testing which techniques transfer from language modeling.
The current record at the time of writing reduces pretraining wallclock time from our baseline of 74.32 minutes to 0.92 minutes on a single GPU (NVIDIA L40S), an 81$\times$ speedup, while matching the baseline predictive performance on subsampled TabArena datasets~\citep{erickson2025tabarena}.
The contribution is not a new tabular model family, rather it is a fixed and reproducible speedrun protocol and a public sequence of records that isolates which practical pretraining optimizations transfer to TFMs.
The remainder of this paper details the rules, evaluation, and the techniques behind the records.

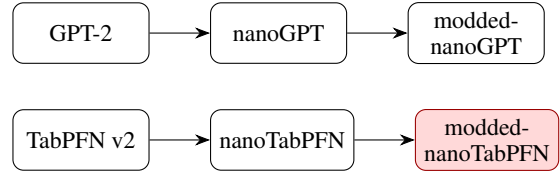
\begin{figure}[t]
  \centering
  \begin{tikzpicture}[
      node distance=6mm and 8mm,
      box/.style={draw, rounded corners, minimum height=8mm, minimum width=18mm, align=center, font=\small},
      ours/.style={box, fill=red!15, draw=red!60!black},
      arr/.style={-{Stealth[length=2mm]}},
    ]
    \node[box]                  (gpt)  {GPT-2};
    \node[box, right=of gpt]    (nano) {nanoGPT};
    \node[box, right=of nano]   (mod)  {modded-\\nanoGPT};
    \node[box, below=of gpt]    (tab)  {TabPFN v2};
    \node[box, right=of tab]    (ntab) {nanoTabPFN};
    \node[ours, right=of ntab]  (mtab) {modded-\\nanoTabPFN};
    \draw[arr] (gpt)  -- (nano);
    \draw[arr] (nano) -- (mod);
    \draw[arr] (tab)  -- (ntab);
    \draw[arr] (ntab) -- (mtab);
  \end{tikzpicture}
  \caption{Two parallel speedrun lineages. The language side (top) is established. modded-nanoTabPFN (red) is the contribution of this paper.}
  \label{fig:lineage}
\end{figure}

\section{Related Work}

\paragraph{Tabular Foundation Models.}
Tabular foundation models pretrain a transformer, typically on synthetic datasets, and then make predictions on new tasks via in-context learning.
TabPFN~\citep{hollmann2023tabpfn,hollmann2025tabpfnv2,grinsztajn2026tabpfn25advancingstateart} pioneered this approach,
followed by TabICL~\citep{qu2025tabicl,qu2026tabiclv2}, TabDPT~\citep{ma2025tabdpt}, and LimiX~\citep{zhang2025limix}.
All these models share an expensive pretraining stage which limits fast prototyping of research ideas.

\paragraph{nanoTabPFN.}
nanoTabPFN~\citep{pfefferle2025nanotabpfn} is a minimal, educational reimplementation of TabPFN v2 and the basis for our competition.
While nanoTabPFN's fast pretraining came from training and evaluating on very small datasets, our baseline scales up the  architecture and the size of the datasets we pretrain on.
We also significantly increase the size limits of the datasets we evaluate on ($5\times$ more datapoints and $10\times$ more features), making it harder to reach the target performance and thereby significantly increasing the baseline training time.
Additionally we also evaluate on all classification tasks of TabArena by subsampling both features and datapoints if they are outside limits, where as nanoTabPFN originally subsampled datapoints but filtered out datasets that exceeded the feature limit.
The most important difference between nanoTabPFN and modded-nanoTabPFN is that the former focused on having very simple and easy-to-understand code for educational purposes, where as in modded-nanoTabPFN we are interested in any techniques that speed up training, even if they require complex, hard-to-understand implementations.

\paragraph{nanoGPT and modded-nanogpt.}
Our format is inspired by community speedrun efforts on language model pretraining.
nanoGPT~\citep{karpathy2023nanogpt} is a minimal, hackable reimplementation of GPT-2 designed for easy prototyping and experimentation.
modded-nanogpt~\citep{moddednanogpt} is a community-driven speedrun built on top of nanoGPT in which contributors compete to reach a fixed validation loss in less wallclock time.
Iterations of the competition have surfaced techniques such as the Muon optimizer~\citep{jordan2024muon}.
We adopt the same speedrun leaderboard format for tabular foundation models.

\section{Competition}
Our competition is an open community speedrun, in which contributors compete to reach a fixed downstream-accuracy target on fixed hardware in the shortest wallclock time.
Following modded-nanogpt, contributors modify a single training script, which logs its own source code, configuration, software versions, GPU metadata, peak memory, and timing information so every record can be reconstructed from the submitted logs.
The official wallclock budget is cumulative training time only, excluding evaluation and prior generation.

\paragraph{Goal.}
The goal is to pretrain a neural network that beats the average validation ROC AUC of a Random Forest baseline on subsampled TabArena datasets, using 1 NVIDIA L40S, in the shortest wallclock time. The target is derived from the same evaluation pipeline as the speedrun model (Appendix~\ref{app:rf-target}).

\paragraph{Rules.}
The rules are minimal and exist only to keep records comparable.
A new record must not change the evaluation pipeline, must not load pretrained weights, and must run faster than the prior record when re-run on the verifier's hardware at the same seed.
To suppress cluster noise, the reported wallclock is the median of multiple verification runs (Appendix~\ref{app:records}), and each submitted log carries the metadata needed to reconstruct the run (Appendix~\ref{app:repro-metadata}).
Beyond that, anything is fair game.

\paragraph{Evaluation.}
Evaluation is on the 38 classification tasks of TabArena~\citep{erickson2025tabarena}, a benchmark for tabular machine learning that curates a representative collection of classification and regression tasks.
We do not adopt the full benchmark protocol but use the curated datasets only, subsampled per task to at most 100 features and 1000 rows for fast iteration.
This lets us inherit the dataset diversity while keeping evaluation fast enough for rapid iteration.
For each task we run 5-fold stratified cross-validation with per-fold preprocessing fitted on train only, concatenate out-of-fold predictions across folds, and score once with ROC AUC, binary for two-class tasks and one-vs-rest for multi-class.
We average the per-task scores over the 38 tasks and run this evaluation periodically during training, stopping as soon as the  target  (performance equivalent to a random forest) is reached. The full pipeline and the task list are in Appendices~\ref{app:eval} and~\ref{app:tabarena-tasks}.

\begin{table}[t]
  \centering
  \small
  \begin{tabular}{@{}rrrl@{}}
    \toprule
    \# & Time (min) & Datasets & Technique \\
    \midrule
    1 & $74.32$ & $80{,}576$ & Baseline \\
    2 & $54.41$ & $45{,}824$ & Muon optimizer \\
    3 & $10.10$ & $13{,}184$ & SDPA, bf16, LR, width \\
    4 & $\phantom{0}9.26$ & $13{,}184$ & Batched Muon, compile \\
    5 & $\phantom{0}7.57$ & $11{,}200$ & Residual decay \\
    6 & $\phantom{0}3.88$ & $\phantom{0}9{,}664$ & RMSNorm, Thinking Rows \\
    7 & $\phantom{0}3.48$ & $\phantom{0}8{,}768$ & LAWA, AdamW WD \\
    8 & $\phantom{0}2.15$ & $\phantom{0}4{,}992$ & Repeated feature grouping \\
    9 & $\phantom{0}0.92$ & $\phantom{0}3{,}648$ & HPO, Muon WD, Mean-pool \\
    \bottomrule
  \end{tabular}
  \caption{Speedrun records table}
  \label{tab:records}
\end{table}

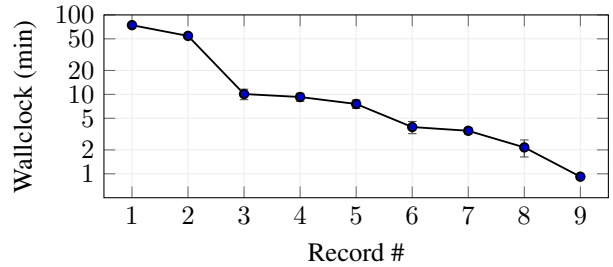
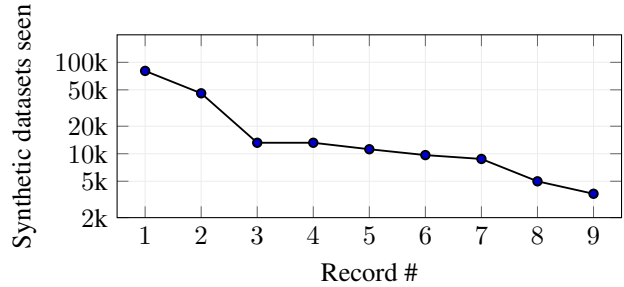
\begin{figure}[t]
  \centering
  \begin{subfigure}{\columnwidth}
    \centering
    \begin{tikzpicture}
      \begin{axis}[
        width=\columnwidth, height=4.0cm,
        xlabel={Record \#}, ylabel={Wallclock (min)},
        ymode=log, log basis y={10},
        xtick={1,2,3,4,5,6,7,8,9},
        xmin=0.5, xmax=9.5, ymin=0.5, ymax=100,
        ytick={1,2,5,10,20,50,100},
        yticklabels={$1$,$2$,$5$,$10$,$20$,$50$,$100$},
        grid=both, grid style={lightgray!30, line width=0.3pt},
        tick style={black!60},
        every axis plot/.append style={mark=*, mark size=1.6pt, line width=0.7pt},
      ]
        \addplot+[
          color=black,
          error bars/.cd, y dir=both, y explicit,
          error bar style={line width=0.5pt, black!70},
        ] coordinates {
          (1, 74.32) +- (0, 0.90)
          (2, 54.41) +- (0, 0.29)
          (3, 10.10) +- (0, 1.52)
          (4,  9.26) +- (0, 1.07)
          (5,  7.57) +- (0, 0.93)
          (6,  3.88) +- (0, 0.68)
          (7,  3.48) +- (0, 0.30)
          (8,  2.15) +- (0, 0.52)
          (9,  0.92) +- (0, 0.04)
        };
      \end{axis}
    \end{tikzpicture}
    \caption{Wallclock to target.}
    \label{fig:ledger-wallclock}
  \end{subfigure}

  \vspace{0.3em}

  \begin{subfigure}{\columnwidth}
    \centering
    \begin{tikzpicture}
      \begin{axis}[
        width=\columnwidth, height=4.0cm,
        xlabel={Record \#}, ylabel={Synthetic datasets seen},
        ymode=log, log basis y={10},
        xtick={1,2,3,4,5,6,7,8,9},
        xmin=0.5, xmax=9.5, ymin=2000, ymax=200000,
        ytick={2000,5000,10000,20000,50000,100000},
        yticklabels={$2$k,$5$k,$10$k,$20$k,$50$k,$100$k},
        grid=both, grid style={lightgray!30, line width=0.3pt},
        tick style={black!60},
        every axis plot/.append style={mark=*, mark size=1.6pt, line width=0.7pt},
      ]
        \addplot+[color=black] coordinates {
          (1, 80576) (2, 45824) (3, 13184) (4, 13184)
          (5, 11200) (6,  9664) (7,  8768) (8,  4992)
          (9,  3648)
        };
      \end{axis}
    \end{tikzpicture}
    \caption{Synthetic datasets to target.}
    \label{fig:ledger-datasets}
  \end{subfigure}

  \caption{Records of Table~\ref{tab:records} on log scale.}
  \label{fig:ledger-plot}
\end{figure}

\begin{figure*}[t]
  \centering
  \includegraphics[width=\textwidth]{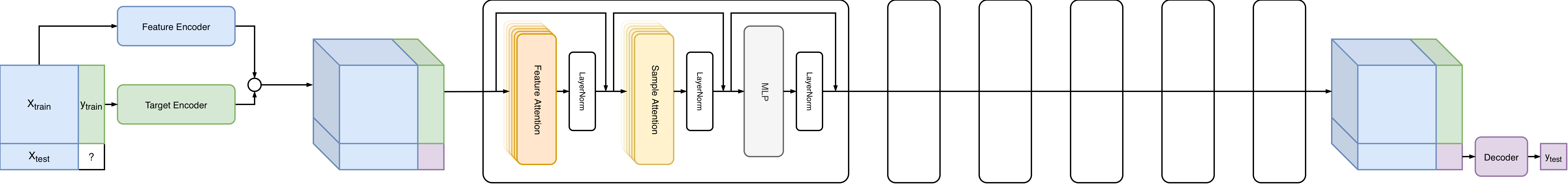}
  \caption{High-level architecture of the baseline. Input features $x$ are encoded with a feature encoder, target labels $y$ with a target encoder and extended with the train-row mean to match the data shape. The two streams are merged and passed through a stack of 6 transformer blocks, each applying feature-axis attention, norm, sample-axis attention, norm, MLP, norm. The test-row outputs are sliced out and fed into the decoder.}
  \label{fig:base}
\end{figure*}

\begin{figure*}[t]
  \centering
  \includegraphics[width=\textwidth]{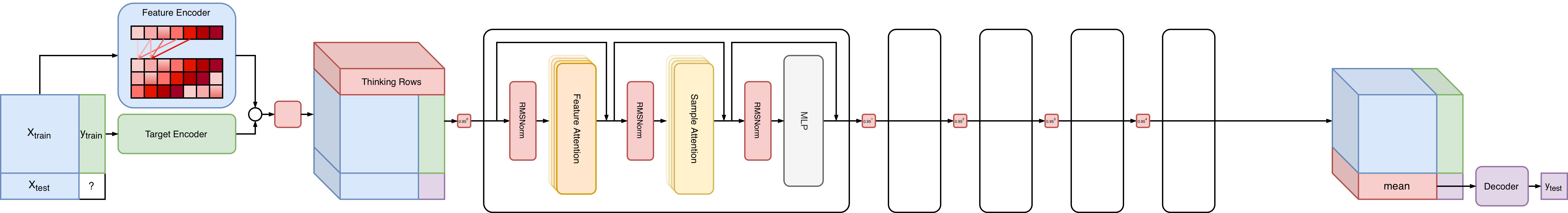}
  \caption{High-level architecture of the current best record. Changes from the baseline are marked in red. Input features are grouped at the feature encoder, 24 learnable thinking rows are appended along the data axis, the residual stream entering each block is decayed by $0.95^i$, the norms move ahead of attention and MLP, and the stack is reduced to 5 blocks. The mean over feature embeddings at the test rows is fed into the decoder instead of the target slice.}
  \label{fig:modded}
\end{figure*}

\section{Speedrun Results}
The records collected so far compress pretraining from $74.32$ minutes on the baseline to $0.92$ minutes on the current best, an $81\times$ wallclock speedup.
The synthetic dataset count drops from $80{,}576$ to $3{,}648$ over the same interval, as summarized in Table~\ref{tab:records} and Figure~\ref{fig:ledger-plot}.
Each subsection below walks through one record, naming the techniques it introduced and its resulting wallclock time.
Each subsequent record builds on the previous one.
See Appendix~\ref{app:configs} for a side-by-side configuration of the baseline against the current best, and Appendix~\ref{app:records} for per-record details and statistics.

\subsection{Baseline}
The baseline is nanoTabPFN consolidated into a single training script:
a 6-layer transformer encoder with 6 heads,
embedding size 192,
and MLP hidden size 768,
using post-norm LayerNorms~\citep{ba2016layernorm}
and trained with schedule-free AdamW~\citep{defazio2024schedulefree},
batch size 1,
with evaluation every 64 steps,
in fp32 throughout.
The high-level inference architecture is shown in Figure~\ref{fig:base}.
The pretraining prior is a static dump of synthetic classification datasets, each with $1{,}000$ rows, up to $20$ features, and up to $8$ classes, generated by the TabICL prior~\citep{qu2025tabicl}.
It reaches the target after $80{,}576$ synthetic datasets and $74.32$ minutes on a single L40S, anchoring the start of the speedrun.

\subsection{Muon Optimizer}
The first record uses the Muon optimizer~\citep{jordan2024muon} for the 2D weight matrices of the transformer encoder, while keeping the rest with the baseline's schedule-free AdamW.
Previously validated in image classification and language-model pretraining, Muon transfers to tabular foundation models with the same hyperparameters as in modded-nanogpt and a Muon learning rate set to $0.1\times$ the schedule-free AdamW learning rate.
Wallclock time drops from $74.32$ to $54.41$ minutes and the synthetic dataset count from $80{,}576$ to $45{,}824$, showing sample efficiency on tabular pretraining as well.

\subsection{SDPA, bf16, Learning Rate, Width}
The next record bundles a scaled dot-product attention (SDPA) rewrite, a switch from post- to pre-Norm in the transformer blocks, and mixed precision training with hyperparameter tunings: a learning rate increase to $10^{-3}$ and a wider embedding ($192\to 256$) with fewer heads ($6\to 4$).
Ablations isolate the learning-rate increase and the SDPA rewrite as the dominant drivers, with bf16 and TF32 matmul contributing the remainder.
Wallclock time drops from $54.41$ to $10.10$ minutes and the synthetic dataset count from $45{,}824$ to $13{,}184$ (per-component ablation in Appendix~\ref{app:records}).

\subsection{Batched Muon, Compile}
The next record batches the Newton--Schulz iteration across the QKV weight matrices and compiles the transformer encoder layer's forward pass.
Both are pure throughput optimizations: wallclock drops from $10.10$ to $9.26$ minutes while the synthetic dataset count is unchanged at $13{,}184$.

\subsection{Residual Decay}
The next record scales the input to each transformer block by an exponentially-decaying factor, with layer $i$'s input scaled by $0.95^i$, exponentially down-weighting earlier-layer contributions in the final output.
Wallclock drops from $9.26$ to $7.57$ minutes and the synthetic dataset count from $13{,}184$ to $11{,}200$.

\subsection{RMSNorm, Thinking Rows}
The next record replaces LayerNorms with lower-precision RMSNorm~\citep{zhang2019rmsnorm} and prepends 16 learnable thinking rows~\citep{grinsztajn2026tabpfn25advancingstateart} to the input.
The data rows attend to these alongside each other in row-axis attention, and the thinking rows account for most of the improvement.
The trained attention maps confirm the thinking rows are actively attended to.
Per-component ablation and the attention maps are in Appendix~\ref{app:records-rmsthink}.
Wallclock drops from $7.57$ to $3.88$ minutes and the synthetic dataset count from $11{,}200$ to $9{,}664$.

\subsection{LAWA, AdamW Weight Decay}
The next record adds Latest Weight Averaging~\citep{kaddour2022lawa,sanyal2024early}, which maintains a FIFO of the last 10 checkpoints and averages them into a temporary model for evaluation only.
Also a weight decay of $0.01$ is applied to the schedule-free AdamW parameters.
Wallclock drops from $3.88$ to $3.48$ minutes and the synthetic dataset count from $9{,}664$ to $8{,}768$.

\subsection{Repeated Feature Grouping}
The next record adopts repeated feature grouping from TabICLv2~\citep{qu2026tabiclv2}: column $j$ is embedded jointly with columns $j{+}1$ and $j{+}3$ (mod $m$), so every column participates in three overlapping groups, addressing representation collapse when features share similar distributions.
Wallclock drops from $3.48$ to $2.15$ minutes and the synthetic dataset count from $8{,}768$ to $4{,}992$.

\subsection{Autoresearch HPO, Muon Weight Decay, Mean-pool Decoder}
The current best record was discovered by adapting autoresearch~\citep{karpathy2026autoresearch}, an LLM-driven hyperparameter and architecture search loop, over multiple runs with human intervention. The changes comprise hyperparameter tuning, weight decay added to Muon, and feeding the decoder the mean over feature embeddings of test rows instead of the target embeddings of test rows.
Wallclock drops from $2.15$ to $0.92$ minutes and the synthetic dataset count from $4{,}992$ to $3{,}648$.

\begin{figure}[t]
  \centering
  \includegraphics[width=\columnwidth]{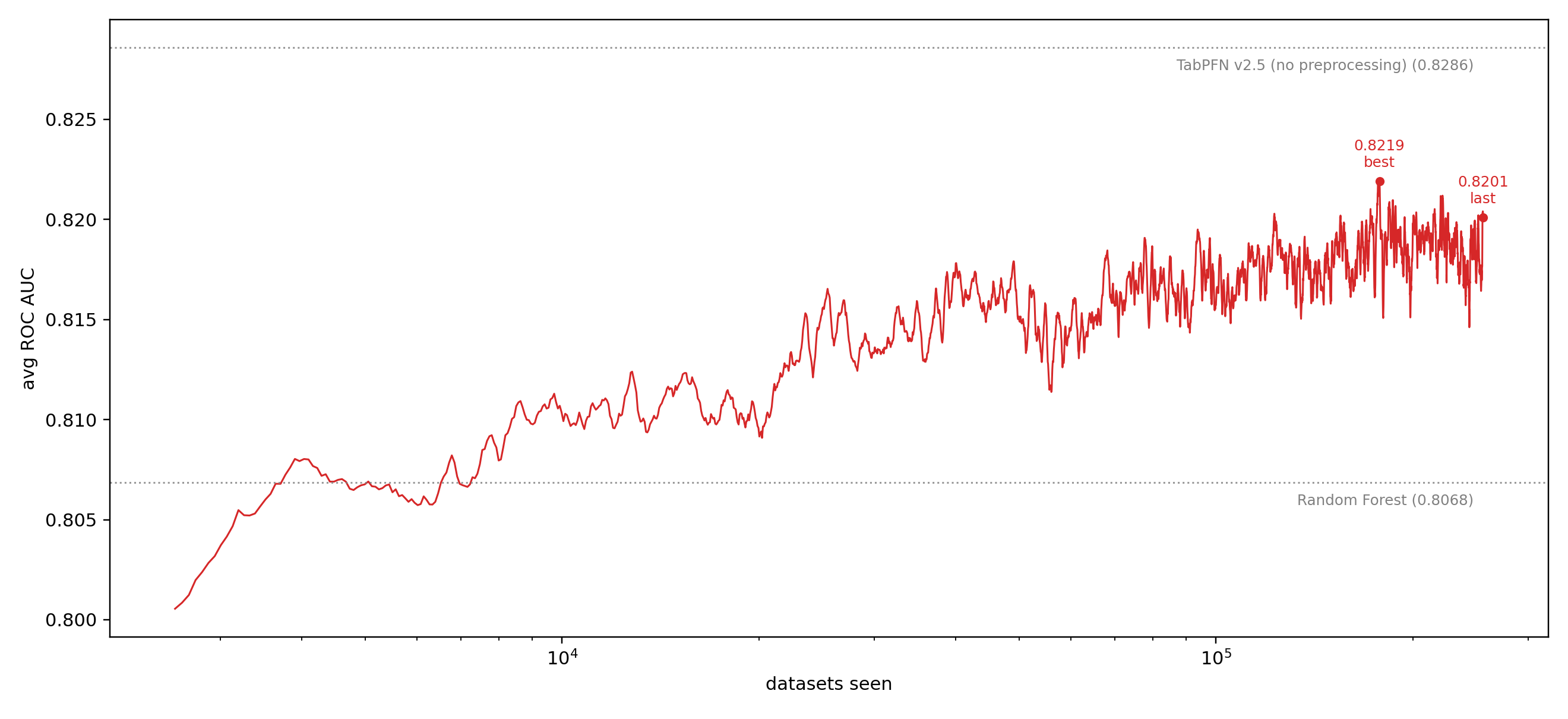}
  \caption{Long-run training trajectory of modded-nanoTabPFN. The x-axis is on a log scale and starts at $2{,}560$ synthetic datasets for visual clarity. Markers indicate the best and last values reached. Dotted lines show the Random Forest target where the speedrun normally stops, and TabPFN v2.5 with no preprocessing as an upper reference.}
  \label{fig:longrun}
\end{figure}

\section{Beyond the Target}
\label{sec:beyond-target}
The high-level inference architecture of the current best record is shown in Figure~\ref{fig:modded}.
While the records optimize pretraining wallclock time to a fixed target, a complementary question is what this model achieves if trained for the baseline's full budget.
Trained until the $256{,}000$-dataset prior dump is exhausted (in $70.88$ minutes), modded-nanoTabPFN reaches an average ROC AUC of approximately $0.82$ on subsampled TabArena, approaching TabPFN v2.5 at $0.8286$ with preprocessing and ensembles disabled (Figure~\ref{fig:longrun}).
At around the same wallclock time the baseline is at $0.8066$ after $76{,}544$ synthetic datasets, essentially still at the Random Forest target, while ours see the entire $256{,}000$-dataset prior in the same wallclock budget. Long-run setup details are in Appendix~\ref{app:longrun}.

\section{Conclusion}
We introduced an open speedrun for tabular foundation model pretraining built around nanoTabPFN, with a fixed downstream target on subsampled TabArena datasets.
The current best record reaches the Random Forest target in $0.92$ minutes on a single L40S, $81\times$ faster than the $74.32$-minute baseline, and using $22\times$ fewer synthetic datasets.
The same recipe scaled to a longer budget keeps improving past the speedrun target, suggesting pretraining ideas selected under tight wallclock also pay off when the budget is relaxed.

Beyond the numbers, the speedrun itself is the contribution, providing a simple protocol for the community to add, verify, and stack pretraining improvements.
Every record so far holds the synthetic prior fixed, making the prior itself the most promising untouched direction.
The leaderboard is open to the tabular foundation model community, and contributions are welcome.

\newpage

\ifdefined\isaccepted
  \section*{Acknowledgements}
  Funded by the European Union. Views and opinions expressed are however those of the author(s) only and do not necessarily reflect those of the European Union or the European Commission. Neither the European Union nor the European Commission can be held responsible for them. This work was supported by the European Union’s Horizon Europe research and innovation programme under grant agreement No 101214398 (ELLIOT).
  \begin{center}
    \begin{minipage}{0.2\textwidth}
      \centering
      \includegraphics[width=\textwidth]{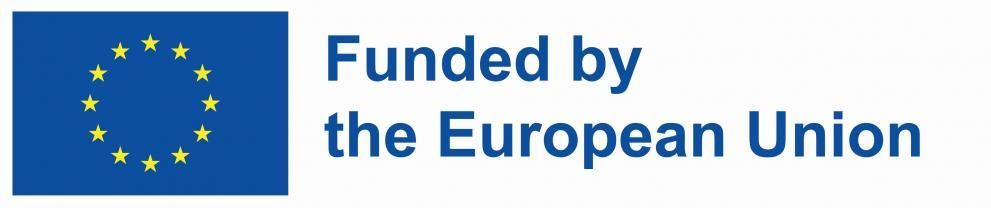}
    \end{minipage}
    \begin{minipage}{0.2\textwidth}
      \centering
      \includegraphics[width=\textwidth]{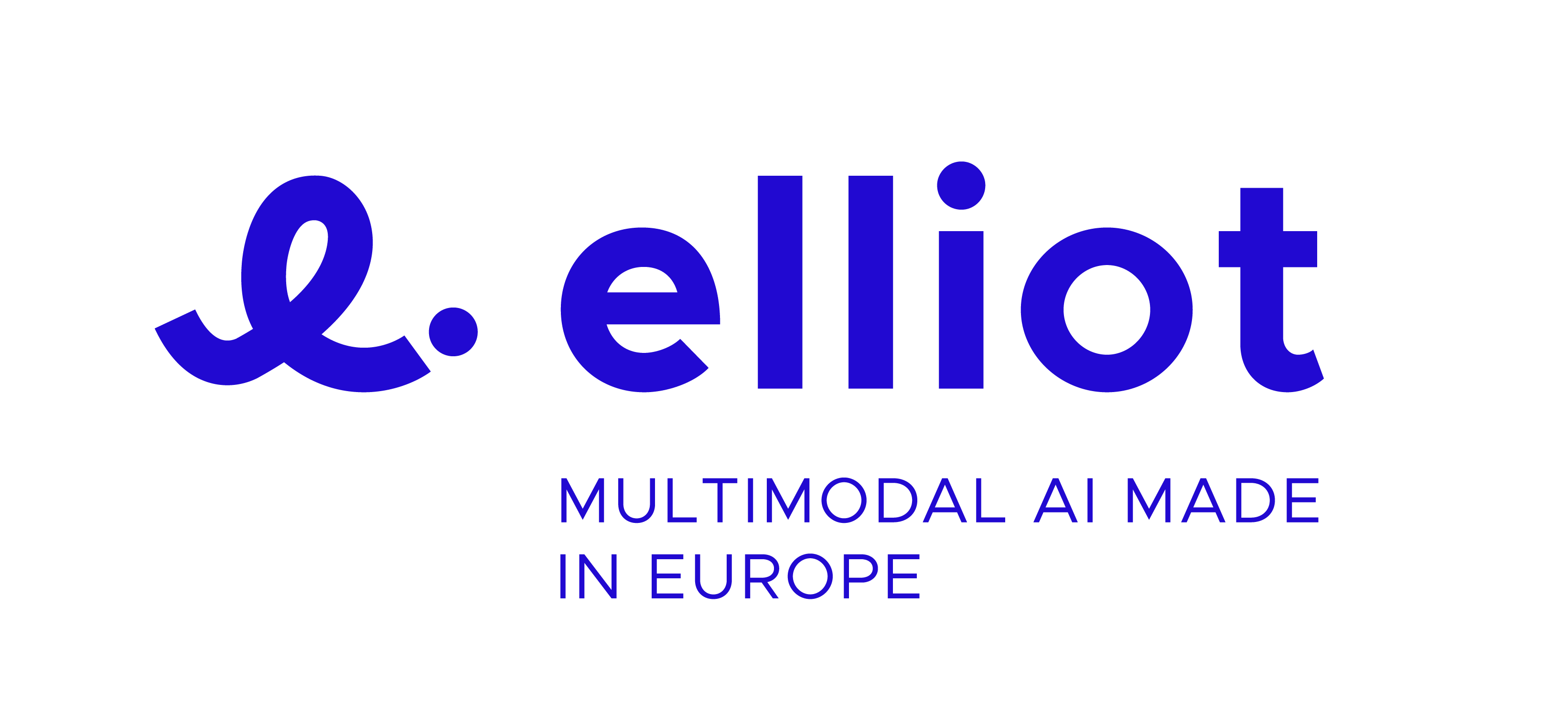}
    \end{minipage}
  \end{center}
  Frank Hutter acknowledges the financial support of the Hector Foundation.
  We thank the reviewers for their feedback and Carter Prince for his contributions to the records.
  We are also grateful to the PyTorch~\citep{paszke2019pytorch} contributors.
\fi

\bibliography{main}
\bibliographystyle{icml2026}

\newpage

\appendix
\onecolumn
\section{Evaluation Details}

\subsection{TabArena Tasks}
\label{app:tabarena-tasks}
Table~\ref{tab:tabarena-tasks} lists the 38 TabArena classification tasks used for evaluation. For subsampling details see Appendix~\ref{app:eval}.

\begin{table}[h]
\centering
\small
\caption{The 38 TabArena classification tasks. Row and feature counts are the original task sizes before subsampling.}
\label{tab:tabarena-tasks}
\begin{tabular}{rrrrl}
  \toprule
  \# & Task ID & Rows & Features & Name \\
  \midrule
  1  & 363613 & $32{,}769$  & $10$      & Amazon\_employee\_access \\
  2  & 363614 & $898$       & $39$      & anneal \\
  3  & 363616 & $76{,}000$  & $171$     & APSFailure \\
  4  & 363618 & $45{,}211$  & $14$      & bank-marketing \\
  5  & 363619 & $10{,}000$  & $11$      & Bank\_Customer\_Churn \\
  6  & 363620 & $3{,}751$   & $1{,}777$ & Bioresponse \\
  7  & 363621 & $748$       & $5$       & blood-transfusion-service-center \\
  8  & 363623 & $5{,}000$   & $20$      & churn \\
  9  & 363624 & $9{,}822$   & $86$      & coil2000\_insurance\_policies \\
  10 & 363626 & $1{,}000$   & $21$      & credit-g \\
  11 & 363627 & $30{,}000$  & $24$      & credit\_card\_clients\_default \\
  12 & 363628 & $129{,}880$ & $22$      & customer\_satisfaction\_in\_airline \\
  13 & 363629 & $768$       & $9$       & diabetes \\
  14 & 363630 & $71{,}518$  & $48$      & Diabetes130US \\
  15 & 363632 & $10{,}999$  & $11$      & E-CommereShippingData \\
  16 & 363671 & $1{,}500$   & $7$       & Fitness\_Club \\
  17 & 363673 & $150{,}000$ & $11$      & GiveMeSomeCredit \\
  18 & 363674 & $2{,}400$   & $31$      & hazelnut-spread-contaminant-detection \\
  19 & 363676 & $10{,}459$  & $24$      & heloc \\
  20 & 363677 & $3{,}845$   & $1{,}618$ & hiva\_agnostic \\
  21 & 363679 & $19{,}158$  & $13$      & HR\_Analytics\_Job\_Change\_of\_Data\_Scientists \\
  22 & 363681 & $12{,}684$  & $25$      & in\_vehicle\_coupon\_recommendation \\
  23 & 363682 & $1{,}723$   & $14$      & Is-this-a-good-customer \\
  24 & 363683 & $50{,}000$  & $213$     & kddcup09\_appetency \\
  25 & 363684 & $2{,}240$   & $26$      & Marketing\_Campaign \\
  26 & 363685 & $1{,}014$   & $7$       & maternal\_health\_risk \\
  27 & 363689 & $7{,}491$   & $87$      & NATICUSdroid \\
  28 & 363691 & $12{,}330$  & $18$      & online\_shoppers\_intention \\
  29 & 363694 & $5{,}910$   & $65$      & polish\_companies\_bankruptcy \\
  30 & 363696 & $1{,}054$   & $42$      & qsar-biodeg \\
  31 & 363699 & $78{,}053$  & $12$      & SDSS17 \\
  32 & 363700 & $2{,}584$   & $16$      & seismic-bumps \\
  33 & 363702 & $3{,}190$   & $61$      & splice \\
  34 & 363704 & $4{,}424$   & $37$      & students\_dropout\_and\_academic\_success \\
  35 & 363706 & $6{,}819$   & $95$      & taiwanese\_bankruptcy\_prediction \\
  36 & 363707 & $1{,}353$   & $10$      & website\_phishing \\
  37 & 363711 & $1{,}699$   & $112$     & MIC \\
  38 & 363712 & $10{,}885$  & $22$      & jm1 \\
  \bottomrule
\end{tabular}
\end{table}

\subsection{Evaluation Pipeline}
\label{app:eval}

\begin{itemize}
  \item \textbf{Subsampling.} If a task has more than 100 features, 100 are selected uniformly at random. If a task has more than 1000 samples, 1000 are selected stratified by class label. Each task is subsampled with a fixed random seed for reproducibility.
  \item \textbf{Cross-validation.} 5-fold \texttt{StratifiedKFold} with shuffling. Class labels are integer-encoded per fold by a \texttt{LabelEncoder} fitted on the training labels.
  \item \textbf{Preprocessing.} Applied per fold, fitted on training data only:
    \begin{itemize}
      \item Columns with at most one unique non-NaN value are dropped.
      \item Numeric columns: \texttt{pd.to\_numeric} coercion followed by mean imputation.
      \item Categorical columns: ordinal encoding (unknown values map to NaN) followed by most-frequent imputation.
    \end{itemize}
  \item \textbf{Metric.} For each task, out-of-fold predictions are concatenated across the 5 folds and scored once with \texttt{roc\_auc\_score}: binary ROC AUC for two-class tasks and one-vs-rest ROC AUC for multi-class tasks. The reported score is the mean of the 38 per-task ROC AUCs.
\end{itemize}
All sources of randomness (feature subsampling, row subsampling, fold splits) are seeded with the same seed (11).

\subsection{Random Forest Target}
\label{app:rf-target}
The competition's downstream target $0.8068462330697953$ is the average validation ROC AUC of an off-the-shelf scikit-learn \texttt{RandomForestClassifier} run through the same evaluation pipeline as the speedrun model.
The number is hardcoded as the \texttt{jackpot} in the training script so every record runs against the same target.

\section{Record Details}
\label{app:records}

The training script logs its own source code, so each record is captured by a single log file.
We verify each record with multiple cluster runs to determine its reported time.
Cluster hosts vary in speed, and the changes from Record 3 onward make training numerically non-deterministic, so both wallclock pretraining time and dataset count vary across runs.
We report the median, which is robust to slow node outliers and selects the verification run whose log is submitted as the record.
The baseline (Record 1) is the only exception: we report its mean, since it sets the reference time the rest of the records are compared against.
Per-record verification statistics are summarized in Table~\ref{tab:record-stats}; subsections below add details specific to individual records.

\begin{table}[h]
\centering
\small
\caption{Verification statistics per record. Mean, standard deviation, and median are wallclock pretraining times in minutes. ``Datasets'' is the synthetic-dataset count at the median run.}
\label{tab:record-stats}
\begin{tabular}{@{}clrrrrr@{}}
  \toprule
  \# & Record & Runs & Mean & Std & Median & Datasets \\
  \midrule
  1 & Baseline                     & $5$  & $74.32$ & $0.90$ & $74.60$ & $80{,}576$ \\
  2 & Muon Optimizer               & $5$  & $54.49$ & $0.29$ & $54.41$ & $45{,}824$ \\
  3 & SDPA, bf16, LR, Width        & $5$  & $10.56$ & $1.53$ & $10.10$ & $13{,}184$ \\
  4 & Batched Muon, Compile        & $5$  & $\phantom{0}9.29$  & $1.07$ & $\phantom{0}9.26$  & $13{,}184$ \\
  5 & Residual Decay               & $31$ & $\phantom{0}7.43$  & $0.93$ & $\phantom{0}7.57$  & $11{,}200$ \\
  6 & RMSNorm, Thinking Rows       & $23$ & $\phantom{0}4.05$  & $0.68$ & $\phantom{0}3.88$  & $\phantom{0}9{,}664$ \\
  7 & LAWA, AdamW Weight Decay     & $27$ & $\phantom{0}3.48$  & $0.30$ & $\phantom{0}3.48$  & $\phantom{0}8{,}768$ \\
  8 & Repeated Feature Grouping    & $31$ & $\phantom{0}2.28$  & $0.52$ & $\phantom{0}2.15$  & $\phantom{0}4{,}992$ \\
  9 & Autoresearch HPO, Muon weight decay, Mean-pool Decoder & $31$ & $\phantom{0}0.93$  & $0.04$ & $\phantom{0}0.92$  & $\phantom{0}3{,}648$ \\
  \bottomrule
\end{tabular}
\end{table}

\subsection{Baseline}
\label{app:records-baseline}
The baseline is the nanoTabPFN-style architecture consolidated into the single-file speedrun script.
It uses the static $256{,}000$-dataset TabICL-prior dump but stops as soon as the Random Forest target is met.
Evaluation is run every 64 optimizer steps, and the official reported time excludes the evaluation calls themselves.
This record is treated as the reference point rather than as a competitor-submitted improvement; consequently Table~\ref{tab:record-stats} reports its mean wallclock time in the main records table, while also listing the median for completeness.

\subsection{Muon Optimizer}
\label{app:records-muon}
The implementation is adapted from modded-nanogpt~\citep{moddednanogpt}.
Muon is applied only to two-dimensional hidden-layer weight matrices in the transformer encoder, while schedule-free AdamW remains responsible for the remaining parameters.
This record is primarily a sample-efficiency improvement.

\subsection{SDPA, bf16, LR, Width}
\label{app:records-carter}
We isolate each component by (i) adding it alone to the Muon record and (ii) removing it alone from the bundled record. Times are the median of 5 runs.

\begin{center}
\small
\begin{tabular}{@{}lrrrrr@{}}
  \toprule
  Component & \multicolumn{2}{c}{Muon $+$ component} & \multicolumn{2}{c}{Record $-$ component} \\
  \cmidrule(lr){2-3} \cmidrule(lr){4-5}
       & median & $\Delta$ & median & $\Delta$ \\
  \midrule
  Learning rate $10^{-4}\to 10^{-3}$ & $31.52$ & $-42\%$ & $32.62$ & $+223\%$ \\
  Embedding $(192,6)\to(256,4)$       & $45.78$ & $-16\%$ & $15.49$ & $+53\%$  \\
  Explicit-QKV SDPA                   & $38.99$ & $-28\%$ & $22.29$ & $+121\%$ \\
  bf16 autocast (training)            & $49.31$ & $-9\%$  & $10.55$ & $+4\%$   \\
  bf16 autocast (inference)           & $49.13$ & $-10\%$ & $\phantom{0}8.87$ & $-12\%$  \\
  TF32 matmul                         & $45.06$ & $-17\%$ & $\phantom{0}9.49$ & $-6\%$   \\
  \bottomrule
\end{tabular}
\end{center}

The learning-rate increase and the SDPA rewrite both are the main drivers.

\subsection{Batched Muon, Compile}
\label{app:records-batched-muon}
This record is intended as a pure throughput change.
The Newton--Schulz iteration used by Muon is batched across the QKV matrices, reducing optimizer overhead without changing the model's training objective or evaluation interface.
In addition, the transformer encoder layer forward pass is compiled.

\subsection{Residual Decay}
\label{app:records-residual-decay}
A later re-run of the same code over 41 cluster runs reached a median of $6.06$ minutes ($11{,}008$ datasets at the median run), illustrating cluster-noise variance in the record-time number.

\subsection{RMSNorm, Thinking Rows}
\label{app:records-rmsthink}
Both changes are adopted from the TabPFN v2.6 release file. RMSNorm replaces all three LayerNorms in the encoder block while skipping the FP32 autocast upcast that PyTorch's LayerNorm applies by default. Thinking rows, originally introduced in the TabPFN v2.5 model report, prepend 16 learnable embeddings of dimension $e$ along the row dimension. Each embedding is broadcast across all columns, and the rows count as part of the train portion of the input.

We tested each change in isolation:

\begin{center}
\small
\begin{tabular}{@{}lrr@{}}
  \toprule
  Run & Median time (min) & Median datasets \\
  \midrule
  RMSNorm alone (11 runs)        & $6.32$ & $11{,}264$ \\
  Thinking rows alone (9 runs)   & $4.22$ & $10{,}304$ \\
  Combined (23 runs)             & $3.88$ & $\phantom{0}9{,}664$ \\
  \bottomrule
\end{tabular}
\end{center}

Thinking rows alone account for most of the combined improvement; RMSNorm contributes a smaller but real additional gain.

\begin{figure}[h]
  \centering
  \includegraphics[width=0.8\linewidth]{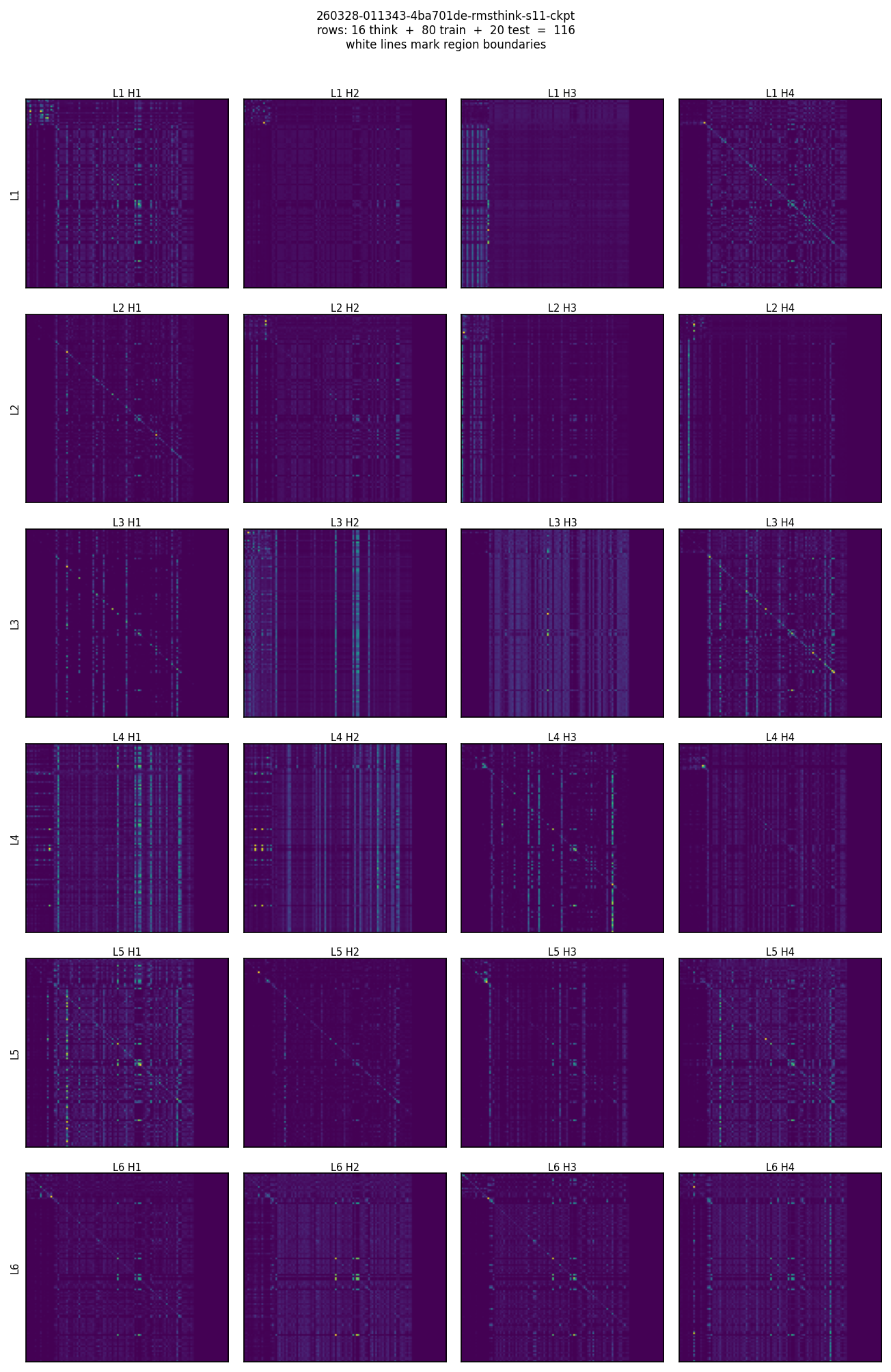}
  \caption{Row attention map of the trained model on the TabArena task \texttt{jm1}, subsampled to 100 datapoints for visual clarity, with all encoder layers and heads shown. The first 16 positions on each axis are the prepended thinking rows. In each panel, rows are queries and columns are keys. Vertical bands at the thinking-token columns show data rows placing attention on the thinking tokens, with layer~1 head~3 in particular concentrating most of its mass on those columns.}
  \label{fig:thinking-attention}
\end{figure}

\subsection{LAWA, AdamW Weight Decay}
\label{app:records-lawa}
Latest Weight Averaging keeps a FIFO buffer of the last 10 checkpoints.
At evaluation time only, the buffered weights are averaged into a temporary model and scored, the training weights themselves continue from the unaveraged trajectory.
The record also adds weight decay $0.01$ to the schedule-free AdamW parameter group.

\subsection{Repeated Feature Grouping}
\label{app:records-featuregroup}
For a table with $m$ columns, the $j$-th group contains columns at positions $(j, j+1, j+3) \bmod m$.
The shift pattern $(0, 1, 3)$ is chosen so that the differences between any two of its elements are all distinct: $1$, $2$, $3$.
This guarantees that for any table with $\geq 7$ columns, no pair of columns co-occurs in more than one group.
The implementation is adapted from nanotabicl.

\subsection{Autoresearch}
\label{app:records-autoresearch}
The hyperparameter changes adopted from the autoresearch search loop are listed in Table~\ref{tab:autoresearch-hpo}.
In addition, weight decay is added to the Muon parameter update, and the decoder is fed the mean over feature tokens at the test rows instead of the target token.
We do not claim that every component in this bundle is independently beneficial, a full leave-one-out ablation of the final record is future work.

\begin{table}[h]
\centering
\small
\caption{Hyperparameter changes adopted from the autoresearch runs.}
\label{tab:autoresearch-hpo}
\begin{tabular}{@{}lrr@{}}
  \toprule
  Hyperparameter & Before & After \\
  \midrule
  \texttt{batch\_size}         & $1$    & $2$    \\
  \texttt{steps}               & $64$   & $32$   \\
  \texttt{l} (encoder layers)  & $6$    & $5$    \\
  \texttt{thinking\_rows}      & $16$   & $24$   \\
  \texttt{feature\_group\_size}& $3$    & $5$    \\
  \texttt{muon\_momentum}      & $0.95$ & $0.96$ \\
  \texttt{grad\_clip}          & $1.0$  & $2.0$  \\
  \bottomrule
\end{tabular}
\end{table}

\section{Baseline and Current Best Configuration}
\label{app:configs}

Table~\ref{tab:configs} summarizes the architecture and training settings of the baseline (Record 1) against those of the current best (Record 9).
Empty cells indicate the option does not exist in that configuration.

\begin{table}[h]
\centering
\small
\caption{Architecture and training configuration of the baseline against the current best record.}
\label{tab:configs}
\begin{tabular}{@{}lll@{}}
  \toprule
  & Baseline & Current Best \\
  \midrule
  \multicolumn{3}{l}{\textit{Architecture}} \\
  Layers ($l$)              & $6$              & $5$ \\
  Heads ($a$)               & $6$              & $4$ \\
  Embedding ($e$)           & $192$            & $256$ \\
  MLP hidden ($h$)          & $768$            & $768$ \\
  Norm                      & post-LayerNorm   & pre-LowerPrecisionRMSNorm \\
  Feature group size        & $1$              & $5$ \\
  Thinking rows             & $0$              & $24$ \\
  Residual decay            & $1.0$            & $0.95$ \\
  Decoder input             & target token     & mean of feature tokens \\
  \midrule
  \multicolumn{3}{l}{\textit{Optimization}} \\
  Optimizer                 & schedule-free AdamW & schedule-free AdamW $+$ Muon \\
  Learning rate             & $10^{-4}$        & $10^{-3}$ \\
  AdamW weight decay        & $0$              & $0.01$ \\
  Muon learning rate        & --               & $0.1\times$ AdamW \\
  Muon momentum             & --               & $0.96$ \\
  Muon weight decay         & --               & $0.1$ \\
  Gradient clip             & $1.0$            & $2.0$ \\
  Latest weight averaging   & --               & $K{=}10$\\
  \midrule
  \multicolumn{3}{l}{\textit{Throughput}} \\
  Batch size                & $1$              & $2$ \\
  Training precision        & fp32             & bf16 autocast \\
  \bottomrule
\end{tabular}
\end{table}

\section{Long-Run Experiment}
\label{app:longrun}
The long-run training trajectory in Figure~\ref{fig:longrun} is produced by running the current best record's configuration against the same prior dump but with the target effectively disabled so the run does not exit on a target hit.
The run was carried out on a single L40S and finished in $4252.72$ seconds ($70.88$ minutes); the highest evaluated mean ROC AUC was $0.8219$, with the per-epoch trajectory plotted in Figure~\ref{fig:longrun}.

The TabPFN v2.5 reference $0.8286$ in the same figure is the score of \texttt{TabPFNClassifier} with preprocessing and ensembles disabled (\texttt{PreprocessorConfig(name="none")} and \texttt{n\_estimators}{=}1) run through the identical evaluation pipeline of Appendix~\ref{app:eval} on the same 38 tasks; preprocessing and ensembling are disabled to compare model capacity at the same evaluation interface.

\section{Reproducibility Metadata}
\label{app:repro-metadata}
Each submitted record log contains the complete training script, command-line configuration, random seed, target value, evaluation, measured training wallclock, peak CUDA memory, and the host/GPU/software metadata printed at the beginning and end of the run.
For hardware comparability, records are verified on a single NVIDIA L40S GPU and are re-run by the verifier rather than accepted from submitter-reported timings.

\section{Use of Generative AI Tools}
\label{app:genai-usage}
LLMs were used to set up this speedrun and to iterate on the records, and the current best record was discovered by running autoresearch agents, an LLM-driven agentic loop.
Generative AI tools were also used to assist in drafting and editing this manuscript, with all technical content, claims, and experimental results authored and verified by the human authors, who take full responsibility for the contents.

\end{document}